\documentclass{article}
\usepackage{spconf,amsmath,graphicx}

\usepackage{latexsym}
\usepackage{url}
\usepackage{hyperref}
\usepackage{amsthm}
\usepackage{amssymb}
\usepackage{graphicx}
\usepackage{booktabs}
\usepackage{placeins}
\usepackage{multirow}
\usepackage{tabularx}
\usepackage{epstopdf}
\usepackage{color}
\usepackage{amsfonts}
\usepackage{amsmath}
\usepackage{balance}
\usepackage{enumitem}
\usepackage{xcolor}
\usepackage{wrapfig}
\usepackage{graphics}
\usepackage{todonotes}
\usepackage{makecell}
\usepackage{bbm}
\usepackage{bm}
\usepackage{rotating}
\usepackage{pifont}
\newcommand{\cmark}{\ding{51}}%
\newcommand{\xmark}{\ding{55}}%

\title{Multimodal Modeling for Spoken Language Identification}
\name{
\begin{tabular}{c}
    \it Shikhar Bharadwaj${}^{*}$, Min Ma${}^{*}$, Shikhar Vashishth${}^{*}$\thanks{${}^{*}$Equal Contributions.}, Ankur Bapna, \\ 
    \it Sriram Ganapathy, Vera Axelrod, Siddharth Dalmia, Wei Han, Yu Zhang, \\ 
    \it Daan van Esch, Sandy Ritchie, Partha Talukdar${}^{\dagger}$, Jason Riesa${}^{\dagger}$\thanks{ ${}^{\dagger}$Equal Advising Contributions.}
\end{tabular}
\vspace{-0.5em}
}
\address{Google \\ 
\small{\texttt{\{shikharop, minm, shikharv, partha, riesa\}@google.com}}}
\begin{document}
\ninept
\maketitle

\newcommand{\refalg}[1]{Algorithm \ref{#1}}
\newcommand{\refeqn}[1]{Equation \ref{#1}}
\newcommand{\reffig}[1]{Figure \ref{#1}}
\newcommand{\reftbl}[1]{Table \ref{#1}}
\newcommand{\refsec}[1]{Section \ref{#1}}

\newcommand{\reminder}[1]{\textbf{\textcolor{blue}{#1}}\typeout{#1}}

\newcommand{\add}[1]{\textcolor{red}{#1}\typeout{#1}}
\newcommand{\remove}[1]{\sout{#1}\typeout{#1}}

\newcommand{\method}{MuSeLI}

\newcommand{\problem}{DD}
\newcommand{\problemfull}{Document Dating}

\newcommand{\mc}[1]{\mathcal{#1}}
\newcommand{\bmm}[1]{\bm{\mathcal{#1}}}
\newcommand{\real}[1]{\mathbb{R}^{#1}}

\newcommand{\tensor}{\mathcal{X}}
\newcommand{\Real}{\mathbb{R}}

\newcommand{\tuples}{\mathbb{T}}

\newcommand{\argmax}{arg\,max}

\newcommand\norm[1]{\left\lVert#1\right\rVert}

\newcommand{\note}[1]{\textcolor{red}{[[#1]]}}

\newcommand*{\Scale}[2][4]{\scalebox{#1}{$#2$}}%
\newcommand*{\Resize}[2]{\resizebox{#1}{!}{$#2$}}%
\definecolor{officegreen}{rgb}{0.0, 0.5, 0.0}
\def\mat#1{\mbox{\bf #1}}

\def\x{{\mathbf x}}
\def\L{{\cal L}}
\begin{abstract}
Spoken language identification refers to the task of automatically predicting the spoken language in a given utterance. 
Conventionally, it is modeled as a speech-based language identification task.
Prior techniques have been constrained to a single modality; however in the case of video data there is a  wealth of other metadata that may be beneficial for  this task.
In this work, we propose \method{}, a \textbf{Mu}ltimodal \textbf{S}pok\textbf{e}n \textbf{L}anguage \textbf{I}dentification method, which delves into the use of various metadata sources to enhance language identification.
Our study reveals that metadata such as video title, description and geographic location provide substantial information to identify the spoken language of the  multimedia recording.
We conduct experiments using two diverse public datasets of YouTube videos, and obtain state-of-the-art results on the language identification task.
We additionally conduct an ablation study that describes the distinct contribution of each modality for language recognition.
\end{abstract}

\section{Introduction}
\label{sec:introduction}

\begin{figure*}[t]
	\centering
	\includegraphics[scale=0.85]{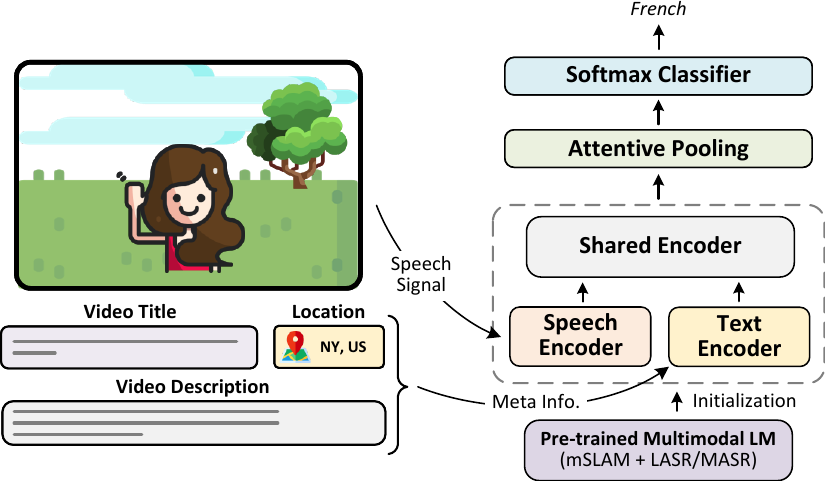}
	\caption{\label{fig:overview}\small Overview of \method{}, a framework to encode both speech and text modalities,  allows to leverage different pre-trained models to initialize speech encoder, text encoder and shared multimodal encoder. Pooling and Softmax layers are added during fine-tuning and randomly initialized. Please see \ref{sec:details} for more details.}
	\vspace{-0.2in}
\end{figure*}

Spoken language identification (LangID) is the task of automatically recognizing the language of a given multimedia recording.
This task serves as a foundational step in the initial stages of multimodal information extraction and analysis.
Precise LangID can aid content recognition, language modeling, and  other downstream tasks such as automatic speech recognition and speech intent understanding \cite{li2013spoken, hou20_interspeech}. 

For multimedia recordings in the wild, such as videos from YouTube, LangID is more challenging due to the presence of multiple speakers, diverse accents and dialects, background non-speech content and noise \cite{sadjadi2018performance}. 
One of the earliest attempts to evaluate this setting is the 2017 NIST language recognition evaluation (LRE) \cite{sadjadi20182017}, where the audio from video \cite{tracey2018vast} was consistently observed to be more challenging \cite{sadjadi2018performance}. Video annotation for speech technologies (VAST) \cite{tracey2018vast} is another common corpus for video LangID.

Most prior efforts in this domain have focused on extracting the spoken content of videos followed by modeling of language classes inherent in the speech data. 
PPRLM \cite{pprlm} created an avenue to generate textual information for spoken langID by using multiple phoneme recognition systems to transcribe unlabeled speech. 
While this research direction remained popular in the past decades, its dependencies on separately trained phoneme recognition systems pose a challenge for spoken langID of low-resource languages, which suffer from limited availability of supervised speech-phoneme data. 

Recently, there has been a growing interest in exploring joint modeling techniques for both speech and text data, aiming to construct a shared encoding space for representations.
Unified speech-text models, such as mSLAM \cite{mslam} and Maestro \cite{maestro} have enabled derivation of speech-text representations that improve downstream tasks such as automatic speech recognition (ASR). 
Text injection for enhancing speech representation learning has also been explored for low-resource ASR tasks \cite{maestro_u}. 
Other related efforts include text-induced losses for speech model pre-training by Tan et al. \cite{tan2023sentence} as well as a student-teacher framework \cite{duquenne2023sentence} for text-based supervision in speech representation learning. 
These endeavors underscore the advantages of having a common embedding space for both speech and text.

In this paper, we present a multimodal framework designed to enhance spoken language recognition by harnessing a wide range of metadata associated with multimedia recordings. 
We term it \textbf{Mu}ltimodal \textbf{S}pok\textbf{e}n \textbf{L}anguage \textbf{I}dentification (\textbf{\method{}}).
In addition to the audio data, multimedia recordings include supplementary metadata such as title, description, geographic location of uploaded videos, etc. These metadata can provide important context for the content embedded in the video recording, and can be especially useful for distinguishing acoustically similar languages.  
We show that the effective use of such information can improve the language recognition performance significantly. Our contributions include:
\begin{itemize}[itemsep=2pt,parsep=2pt,partopsep=2pt,leftmargin=*,topsep=2pt]
   \item We propose a multimodal framework that facilitates the incorporation of diverse metadata associated with a multimedia recording for spoken LangID. It does not depend on separately trained text LangID models, nor on text LangID labels.
   \item To the best of our knowledge, this study is the first attempt to demonstrate that, despite being noisy, video title, description, and geographic location can improve spoken LangID performance.
   \item Our proposed method achieves state-of-the-art performance on public benchmarks. It is also shown to be effective in distinguishing acoustically similar and low-resource languages.
\end{itemize}

\section{Related Works}
\label{sec:related_works}

\textbf{Text LangID} - Previous works used n-gram based techniques \cite{Goldhahn2012BuildingLM,OrtizSuarez2019AsynchronousPF} for this task. Recently, Caswell et al. \cite{wild_langid} have explored text LangID in the context of web-crawl corpora.
The authors trained LangID models for classifying  $1\text{,}629$ languages and explored a variety of methods to mitigate classification errors. When it comes to YouTube (YT) video title and description, there is no gold text langID available, and langID of text in low-resourced languages remains challenging. In this work, we use \emph{unlabeled} text in input, and encourage \method{} to automatically learn how to address the mismatch between language of text and language of audio.

\textbf{Speech LangID} - With the renaissance of deep learning, X-vector \cite{snyder2018spoken} has became status duo for spoken langID.
Time-delay neural networks \cite{snyder2018spoken}, residual networks \cite{miao2021d}, squeeze and excitation models
\cite{valk2021voxlingua107}, and attentive pooling with conformer models \cite{wang2022attentive} have been investigated for more efficient neural networks. 
The LASR \cite{LASR} and MASR \cite{MASR} methods add additional objectives to pre-training for learning language specific representations. 

\textbf{Multimodal LangID} - Multimodal models such as mSLAM \cite{mslam} and Maestro \cite{maestro} were primarily investigated for speech recognition task, and they do not consider textual information of videos. A limited number of methods have explored multimodal modeling for language recognition, but for  music  content analysis \cite{amazon_music_langid,music_mlangid}.

\begin{table*}[!t]
	\centering
	\begin{tabular}{clccccccc}
	    \toprule
	    \multirow{3}{*}{\shortstack{\textbf{Method}}} &
	    \multirow{3}{*}{\shortstack{\textbf{Pre-trained}\\ \textbf{Model}}}  &
	    \multirow{3}{*}{\shortstack{\textbf{Language}\\ \textbf{Aware}}} &
	     \multicolumn{3}{c}{\textbf{Dhwani-YT}} & \multicolumn{3}{c}{\textbf{VoxLingua107}}\\
	    \cmidrule(r){4-6} \cmidrule(r){7-9}
	    & & & Accuracy & F1 & FPR$\downarrow$ &  Accuracy & F1 & FPR$\downarrow$ \\
		\midrule
		\multirow{3}{*}{\shortstack{Speech-only\\ LangID}} & mSLAM & \cmark & 63.6 & 48.4 & 1.8e-2 & 81.9 & 73.9 & 3.2e-3 \\
		
		& mSLAM-YT & \xmark & 64.7 & 50.1 & 1.7e-2  &  92.4 & 91.1 & 5.8e-4 \\
		& mSLAM-YT & \cmark & 66.1 & 51.2 & 1.6e-2  &  93.0 & 91.6 & 6.4e-4 \\
		\midrule
		\multirow{3}{*}{\method{}} & mSLAM  & \cmark & 69.6 & 54.8 & 1.5e-2 & 95.6 & 94.6 & 5.3e-4 \\
		& mSLAM-YT & \xmark & 72.1 & 56.1 & 1.4e-2 & \textbf{96.5} & \textbf{97.1} & \textbf{2.2e-4} \\
		& mSLAM-YT & \cmark & \textbf{72.7} & \textbf{57.6} & \textbf{1.3e-2} & 96.2 & 95.3 & 2.4e-4 \\
		\bottomrule
	\end{tabular}
	\caption{\label{tbl:langid_main} Results on VoxLingua107 and Dhwani datasets. The \method{} variants perform better on both datasets and across all metrics. Please see \refsec{sec:results_main} for details.}
	\vspace{-.1in}
\end{table*}

\section{Method}
\label{sec:details}
In this paper, we propose to learn multimodal representation of speech and text inputs with a unified multimodal framework. A comprehensive overview of our proposed multimodal language recognition system, \method{}, is shown in Figure~\ref{fig:overview}.
\method{} is based on mSLAM \cite{mslam}, which processes speech and text by modality-specific encoders, followed by a multimodal encoder. mSLAM is pre-trained on unsupervised speech and text data using contrastive and masked language modeling objectives \cite{bert}. It also utilizes paired speech-text data through CTC loss \cite{ctc_loss} to learn speech-text alignment. In this work, we enhance an existing pre-trained mSLAM model by incorporating LASR \cite{LASR} pre-training. LASR utilizes language-related metadata to enhance the discriminative capabilities of a speech model with respect to different languages.

\textbf{Multimodal Embeddings} -
In spoken language recognition, a given multimedia recording $\bm{\upsilon}$ comprises of a raw audio waveform $\bm{\mathcal{X}}$ and associated metadata information $\{\phi_1, \phi_2, ... \}$, where $\phi_j$ corresponds to distinct metadata attributes pertaining to $\bm{\upsilon}$. The input audio data $\bm{\mathcal{X}}$ undergoes processing through the speech encoder, which consists of multiple CNN layers followed by a stack of conformer layers \cite{conformer}, to produce latent audio representation $\bm{L}$. 
All metadata information is concatenated to produce a combined text sequence 
\vspace{-0.05in}
\begin{equation}
    \bm{\mathcal{T}} = [\phi_{1}\texttt{[SEP]}\phi_{2}\texttt{[SEP]} ...],
\end{equation}
where $\texttt{[SEP]}$ is a separator tag that allows model to discern between different metadata types. In this work, we utilize three types of metadata: (1) \textit{title}, which is a single sentence summary of the entire recording, (2) \textit{description}, which provides a  detailed explanation of the content, and (3) \textit{upload  location}, which indicates the region and country the recording was uploaded from. While these signals may exhibit noise and lack a direct connection to the identity of the spoken language, we hypothesize that they may lead to enhanced performance on the task. The metadata text sequence $\bm{\mathcal{T}}$ is input to the text encoder, which consists of a token embedding layer to generate the latent representation $\bm{T}$ for the metadata.
Finally, the concatenated speech and metadata  embeddings $[\bm{L}; \bm{T}]$ are passed to the multimodal encoder to produce a unified representation $\bm{H}$ for the entire multimedia recording. 

\textbf{Weighted Layer Representation} - The multimodal encoder consists of a series of conformer layers. Hsu et al. \cite{hubert}  demonstrated that the representations generated by the final layer may not be optimal for all tasks. Hence, we take a weighted combination of representations from all layers where weights are kept learnable and are trained using backpropagation, i.e., 
\begin{equation}\label{eqn:weighted_rep}
    \bm{H} = \sum_k\alpha_k \bm{H}_k,
\end{equation}
where $\bm{H}_k$ denotes the representation from $k^{\text{th}}$ conformer layer of the multimodal encoder and $\alpha_k$ is a learnable parameter corresponding to each layer. The weighted representation provides flexibility to the model for weighing different layers of the encoder stack and eliminates the need to carefully choose the layer.

\textbf{Attentive Pooling} - To facilitate the merging of audio and text information, we employ an attention-based pooling, where the pooling is performed on the sequence dimension.
This layer assigns distinct weights to the hidden sequences from the audio and text components, thereby capturing the significance of each modality effectively.
We use a learnable query vector $\bm{Q}$, with $\bm{H}$ as the key and value sequences respectively in the multi-head attention \cite{transformer}. The final pooled vector $\bm{p}$ is computed as,
\begin{equation}\label{eqn:attention_pooling}
    \bm{p} = \mathrm{MultiHead}(\bm{Q}, \bm{H}, \bm{H}).
\end{equation}
Finally, $\bm{p}$ is passed through a soft-max layer for generating class probabilities. We optimize the model on cross-entropy loss over the language classes.

\section{Experimental Setup}
\label{sec:experiments}

\noindent  \textbf{Datasets}
We experiment on the follwoing public datasets derived from YouTube (YT):

\begin{itemize}[itemsep=2pt,parsep=2pt,partopsep=2pt,leftmargin=*,topsep=2pt]
    \item \textbf{Dhwani-YT}\footnote{\href{https://github.com/AI4Bharat/IndicWav2Vec/tree/main/data_prep_scripts/urls}{https://github.com/AI4Bharat/IndicWav2Vec} last accessed on 14th September, 2023.} We   experiment on the publicly available YT portion of the Dhwani dataset \cite{dataset_in_w2v}.
    Dhwani-YT contains $4$k hours of audio   from $1.9$k YT channels.
    This dataset spans over $22$ south Asian languages, which covers $4$ language families and $14$ writing scripts.
    \item \textbf{VoxLingua107} \cite{valk2021voxlingua107} is a language identification dataset composed of $6.6$k hours of audio from approximately $64$k videos.
    The training dataset spans over $107$ languages, while the evaluation set consists of $1\text{,}609$ samples from $33$ languages.

\end{itemize}

\noindent  \textbf{Baselines}
In our experiments, we use the $600$M mSLAM model, which has undergone pre-training with large volume of raw speech and text data, in addition to paired speech-text datasets \cite{mslam}.
We introduce a modified version of mSLAM, referred to as mSLAM-YT, which is pre-trained using YouTube-based datasets employed in Google-USM \cite{google_usm}. 
Additionally, we create another variant of mSLAM by utilizing LASR pre-training on publicly available datasets \cite{LASR}, which leverages language metadata to make speech models language-aware through a contrastive objective.

\noindent \textbf{Evaluation Metrics:} In order to assess the effectiveness of different LangID models, we conduct comparisons based on accuracy, macro-F1 score, precision, and False Positive Rates (FPR).
As shown in \cite{wild_langid}, FPR is a valuable metric for evaluating the efficacy of a LangID system, specifically for low-resource languages. 

\noindent  \textbf{Implementation Details}
We adopt most of the hyper-parameters from previous works \cite{LASR, MASR, dataset_fleurs}.
We use a batch size of $128$ and trim the text sequence to $400$ tokens for the multimodal model.
The speech sequence is trimmed to $1.6$k frames.
On the VoxLingua107 and Dhwani dataset we fine-tune for $26$k and $30$k steps respectively.
We use the Adam optimizer with a linear rate schedule.

\begin{table}[t]
	\centering
	\begin{tabular}{lc}
	    \toprule
	    \multicolumn{1}{l}{\textbf{Model}} & \multicolumn{1}{l}{\textbf{Accuracy}}  \\
		\midrule
		SpeechBrain \cite{speech-brain} & 93.3\\
		XLS-R \cite{xls_r} & 94.3\\
		MMS (VL) \cite{meta_speech_langid} & 94.7 \\
		MMS (4017 languages) & 93.9 \\
		AmberNet \cite{ambernet} & 95.3\\
		\midrule
		\method{} (weighted-layer) & 96.5\\
		\method{} (best-layer) & \textbf{97.6}\\
		\bottomrule
	\end{tabular}
	\caption{\label{tbl:vox_table} \method{} achieves SOTA performance on VoxLingua107. Please see \refsec{sec:results_main} for details.}
	\vspace{-.25in}
\end{table}
\section{Results}
\label{sec:results}



\subsection{Performance Comparisons}
\label{sec:results_main}
As shown in \reftbl{tbl:langid_main}, \method{} outperforms Speech-only LangID on both datasets, in all evaluation metrics, regardless of the choice of pre-trained models. Specifically, by leveraging multimodal signals, \method{} improved accuracy from $93.0$\% to $96.5$\% on Voxlingua, and from $66.1$\% to $72.7$\% on Dhwani-YT. \method{} achieves state-of-the-art performances (cf. \reftbl{tbl:vox_table}), even without including VoxLingua training data in its pre-training (while \cite{xls_r} and \cite{meta_speech_langid} did).
The previous best performance in the spoken LangID task was achieved by AmberNet \cite{ambernet}, a model suited for practical deployment due to its small size.
While in this paper, we propose \method{} to model both speech and text modalities in a unified framework, and its larger model capacity would be a better fit to learn a generally useful representation for multiple speech tasks.

\begin{figure}[t]
	\centering
	\includegraphics[width=\columnwidth]{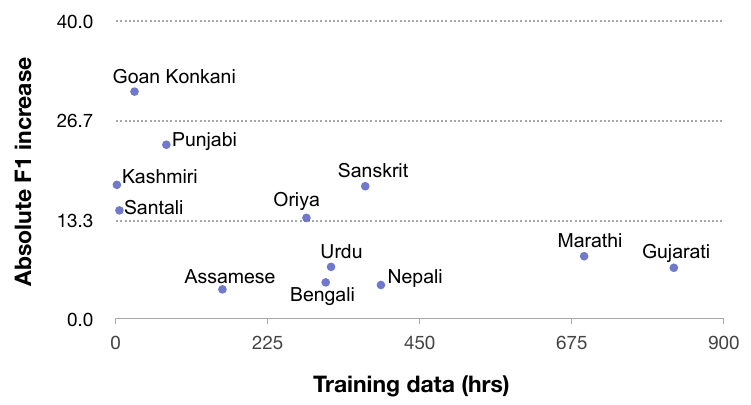}
	\caption{\label{fig:f1_vs_hr}\small Languages with the least amount of fine-tuning data show most improvement on Dhwani-YT. Please see \refsec{sec:results_main} for details.}
	\vspace{-0.1in}
\end{figure}

Dhwani-YT has a larger test set consisting of 35.9k utterances mostly covering low-resource languages. 
\reffig{fig:f1_vs_hr} shows that the biggest improvements were obtained on low-resource languages with the least amount of fine-tuning data.
For Kashmiri, we had only $1.8$ hours of fine-tuning data, yet \method{} is capable to   increase the F1 score from $0.64$\% to $18.7$\%.
Analyzing the distributions of incorrect predictions by the   best performing systems on Dhwani-YT, we find that textual inputs reduce the number of mis-classifications for the most confusing  languages.
For instance, several Northern Indian languages like Punjabi, Santali, and Oriya were mis-classified as Hindi, since Hindi is a high-resourced language from the same region.
With multimodal signals, \method{} alleviates these confusions to achieve significant gains. The most distinctive signal appears to come from the language-specific writing scripts of title and description. 
For example, the scripts for Punjabi (Gurmukhi), Santali (Ol Chiki), Oriya (Oriya) are different from that of Hindi, which uses Devanagari. 
This suggests that \method{} is a simple yet effective way to encode textual side information to enhance spoken LangID performance, especially for low-resourced languages.
Besides, we observed that additional meta-information can reduce the number of confusing languages. For instance, Speech-only model classifies several Marathi recordings as Hindi, Goan Konkani, and Oriya, while for \method{}, Marathi is only mis-classified as Hindi.
Similarly, on VoxLingua107, textual signals help distinguish Urdu from Hindi, and Spanish from Catalan. Given the acoustical similarity and geographical proximity of the two language pairs, title and description again played a significant role: Urdu uses Perso-Arabic script while Hindi uses Devanagari. Further,  grammatical differences may have helped to discriminate Spanish from Catalan.

\begin{table}[t!]
	\centering
	\begin{tabular}{lcc}
	    \toprule
        \textbf{LangID Variant} & \textbf{Dhwani} & \textbf{VoxLingua}\\
		\midrule
		Metadata-only  & 68.3 &  77.0 \\
		Speech-only    & 66.1&  93.0 \\
		\hspace{3mm} + Title and Description   & 68.3 & 93.3 \\
		\hspace{3mm} \hspace{3mm}   + Upload Location    \\
		\hspace{3mm} \hspace{3mm} \hspace{3mm} w/ Mean Pooling & 72.2 & 96.1 \\
        \hspace{3mm} \hspace{3mm} \hspace{3mm} w/ Attentive Pooling & 72.7 & 96.5 \\
		\bottomrule
	\end{tabular}
	\caption{\label{tbl:langid_ablations} Ablation study with language recognition accuracy (\%) for \method{} with different metadata and pooling types. Please see \refsec{sec:results_ablations} for details.}
\end{table}

\begin{figure}[t]
	\centering
	\includegraphics[width=\columnwidth]{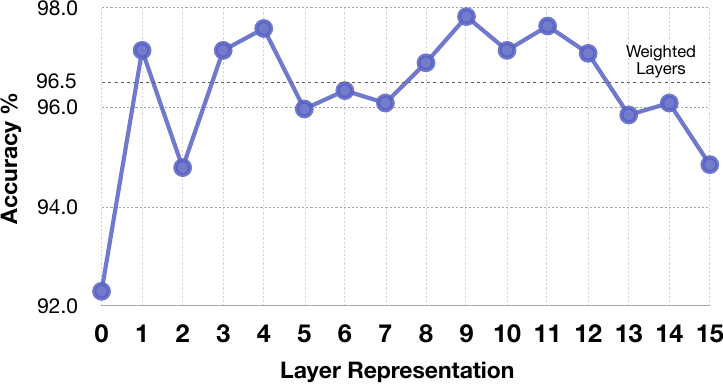}
	\caption{\label{fig:layer_importance}\small LangID performance on using different layer representations for fine-tuning. Please see \refsec{sec:results_layer_importance} for details.}
\end{figure}

\begin{figure}[t]
	\centering
	\includegraphics[width=\columnwidth]{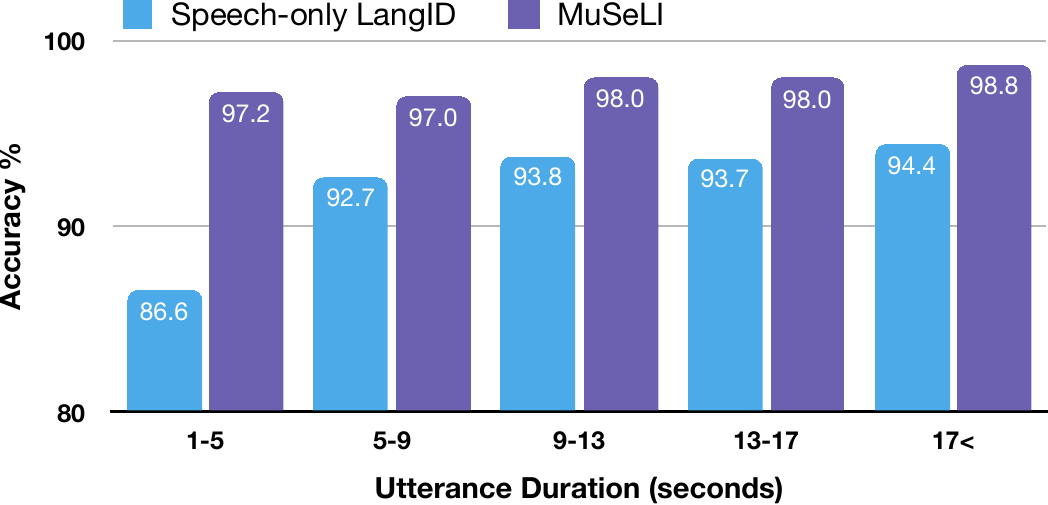}
	\caption{\label{fig:acc_dur_lasr_vox}\small Including meta-information improves accuracy of spoken language identification, across all utterance duration ranges. Please see \refsec{sec:results_duration} for details.}
	\vspace{-.2in}
\end{figure}

\vspace{-0.1in}
\subsection{Effect of different Metadata and Pooling on Performance}
\vspace{-0.05in}
\label{sec:results_ablations}
From the various ablations in \reftbl{tbl:langid_ablations}, we can observe the impact each metadata has on the LangID task.
The upload location is a prominent indicator of the language of the multimedia recording.
However, only adding the title and description can also boost LangID accuracy by a fair margin.
Interestingly, we note that using a metadata only model (containing title, description and upload location) without any speech signal does performs competitively on the Dhwani-YT dataset compared to the baseline Speech-only LangID system.

Results in \reftbl{tbl:langid_ablations} also indicate that attentive pooling (\refeqn{eqn:attention_pooling}) is better than mean pooling in aggregating information over multiple modalities, since it learns to attend to the indicative parts.

\vspace{-0.05in}
\subsection{Estimating Importance of Different Encoder Layers}
\vspace{-0.05in}
\label{sec:results_layer_importance}
The attentive pooling outlined in \refeqn{eqn:attention_pooling}, can be applied over outputs from any layer ($\bm{H}_k$) of the conformer stack.
We fine-tune our best performing mSLAM variant up to the $k$th layer and plot the results in \reffig{fig:layer_importance}.
We observe that the intermediate layers are better than using the last layer for finetuning on the LangID task.
However, we also note that fine-tuning and evaluating all layers of the model is expensive.
On the other hand, our proposed weighted representation scheme (\refeqn{eqn:weighted_rep}) performs comparatively similar to    the best layer representation, while being computationally efficient.
Our best layer selection led to highest accuracy of 97.6\% on Voxlingua107 dataset.

\subsection{Robustness to Utterance Duration}
\vspace{-0.05in}
\label{sec:results_duration}
To investigate the sensitivity of LangID models to utterance duration, we calculated the accuracy over different utterance durations on the Voxlingua107 corpus. 
As illustrated in \reffig{fig:acc_dur_lasr_vox}, both the speech-only and \method{} models generally perform better on longer input utterances. 
More importantly, \method{} is robust to all the utterance duration conditions.  In particular, the largest gain for the use of meta data is seen on the most challenging condition of short duration utterances ($1$-$5$ seconds).
This is expected since meta-information is consistent for the utterances derived from the same video, regardless of the utterance duration.
This analysis highlights that multimodal signals are substantially important when audio signal information is sparse.

\section{Conclusion}
\label{sec:conclusion}
\vspace{-.1in}

We introduce a general multimodal modeling framework, and explore its effectiveness for spoken langID of videos by experimenting on various unlabeled textual metadata information besides speech.
Our proposed method \method{} shows substantial improvements over the speech-only baselines across multiple datasets and different baseline models ($10$\% relative improvement on Dhwani-YT and $4$\% on Voxlingua107).
We conduct comparative studies to show how textual meta-information helps to disentangle similar and low-resourced languages. We also highlight the benefits of utilizing the metadata in short duration audio recordings.

\bibliographystyle{IEEEbib}
\bibliography{mybib.bib}



\end{document}